\documentclass[10pt, conference, compsocconf]{IEEEtran}
\usepackage{gensymb}
\usepackage{subcaption}

\ifCLASSINFOpdf
  \usepackage[pdftex]{graphicx}
\else
\fi
\usepackage{url}

\usepackage{float}

\begin{document}
%
\title{Effect of Rotation Angle in Self-Supervised Pre-training is Dataset-Dependent}


\author{\IEEEauthorblockN{Amy Saranchuk}
\IEEEauthorblockA{Division of Engineering Science\\
University of Toronto\\
Toronto, Ontario, Canada\\
\url{amy.saranchuk@mail.utoronto.ca}}
\and
\IEEEauthorblockN{Michael Guerzhoy}
\IEEEauthorblockA{Division of Engineering Science and\\
Dept. of Mechanical \& Industrial Engineering\\
University of Toronto\\
Toronto, Ontario, Canada\\
\url{guerzhoy@cs.toronto.edu}}
}


%


\maketitle

\begin{abstract}
Self-supervised learning for pre-training (SSP) can help the network learn better low-level features, especially when the size of the training set is small. In contrastive pre-training, the network is pre-trained to distinguish between different versions of the input. For example, the network learns to distinguish pairs (original, rotated) of images where the rotated image was rotated by angle $\theta$ vs. other pairs of images. In this work, we show that, when training using contrastive pre-training in this way, the angle $\theta$ and the dataset interact in interesting ways. We hypothesize, and give some evidence, that, for some datasets, the network can take ``shortcuts" for particular rotation angles $\theta$ based on the distribution of the gradient directions in the input, possibly avoiding learning features other than edges, but our experiments do not seem to support that hypothesis. We demonstrate experiments on three radiology datasets. We compute the saliency map indicating which pixels were important in the SSP process, and compare the saliency map to the ground truth foreground/background segmentation. Our visualizations indicate that the effects of rotation angles in SSP are dataset-dependent. We believe the distribution of gradient orientations may play a role in this, but our experiments so far are inconclusive.

\end{abstract}


%
\IEEEpeerreviewmaketitle

\section{Introduction}

Machine learning has demonstrated remarkable success in various computer vision-for-medical imaging tasks, such as classification and segmentation. However, a major challenge in using these models is the requirement for large labelled datasets, which are costly to create, especially in a field where expert annotation is required. With an abundance of unlabelled data available, self-supervised pre-training is a promising approach learning low-level features without the need for large labelled datasets.

In this paper, we investigate a common self-supervised learning for pre-training method: augmenting the dataset via rotation, and pre-training the network to distinguish between pairs of images where one is rotated by $\theta$ vs other pairs. We observe that the network behaves differently depending on the rotation angle $\theta$, and that this behaviour is dataset-dependent.

We refer to semi-supervised learning for pretaining using a pretext task as ``semi-supervised pretraining" (SSP).

We explore the behaviour of the network by inspecting the correspondence between saliency maps produced by the SSP-trained network using rotation by $\theta$ and the ground-truth segmentation image.

We hypothesize that this has to do with the distribution of the directions of the gradients in the image, and provide some preliminary (but inconclusive) experiments that address this hypothesis.

Work going back to~\cite{lowe1999object} and~\cite{dalal2005histograms} indicates that hisotgrams of oriented gradients (HoGs) often characterize the appearance of objects very well. If the network can rely on HoG-like features to perform the classification task of rotated vs. non-rotated, it would not learn any other features. We hypothesize that this is a reason that some rotation angles seem to produce ``worse" features. We explore this intuition by correlating the performance of SVM-classified HoG features for the unrotated vs. rotated by $\theta$ task, and the correspondence between saliency maps produced by the SSP-trained network using rotation by $\theta$ and the ground-truth segmentation image. We present an experiment to attempt to confirm the hypothesis, but the experiment does not seem to support the hypothesis.



\section{Background}

\subsection{Self-Supervised Pre-training (SSP)}

In SSP, the data ``supervises itself" during training instead of using labels as an indication of its performance by creating artificial supervisory signals from unlabelled data. It allows the network to leverage unlabelled data by learning meaningful representations without manual annotations~\cite{4}~\cite{5}.

Self-supervised pertaining is used when there are two stages: the pretext task and the downstream task. The overall idea behind self-supervised learning is to first pre-train the model using the pretext task and then further fine-tune it for a specific task. The pretext task is used for pre-training and guides the model toward learning intermediate representations of the data. The downstream task is the task which we wish to solve, and to which we wish to transfer knowledge from the pretext task~\cite{2}.

Pretext tasks are unrelated to the downstream task of interest and are used to generate pseudo-labels from data without human annotation. They force the network to learn information about the data such as where edges, colours, and shapes appear in an image. Some common pretext tasks include recolouring an image, predicting the rotation angle, and unscrambling a Jigsaw puzzle~\cite{6}. The downstream task is the primary task of interest (for example, image classification or segmentation).


\subsection{Contrastive Learning}

A popular discriminative pretext approach is contrastive learning. For example, Mikolov et al.~\cite{7} proposed such a method to learn word embeddings. Since then, it has been adapted to other domains and according to \cite{18}, has ``revolutionized" the field of computer vision through learning meaningful representations that can be applied to a variety of vision tasks from unlabelled data.

The general idea behind contrastive learning is that various transformations of the same image still hold the same semantic information~\cite{6}~\cite{15}. The goal is to bring data which is semantically-similar closer together in an embedding space, while pushing dissimilar data apart~\cite{4}~\cite{7}~\cite{15}.

The following outlines the contrastive learning algorithm. First, each sample image from the training set is augmented. The image along with its augmentation are considered a positive pair, and the image along with every other image in the dataset are considered negative pairs. The model is then trained so that it can learn to distinguish between positive and negative pairs by minimizing the distance between positive pairs, and maximizing the distance between negative pairs. By doing so, the model learns quality representations of each image, the knowledge of which is subsequently transferred to the downstream task~\cite{7}. 

Some common data augmentation approaches for creating positive pairs include cropping, flipping, rotating, and adding noise to the image. However, like pretext tasks, not all data augmentations are equally effective. It’s crucial to choose augmentations that preserve the semantic information of the data. 





There have been numerous frameworks proposed to improve vanilla contrastive learning, with one of the most popular being Momentum Contrast (MoCo)~\cite{32}. Proposed in 2020, this method introduces a momentum-encoded queue to retain negative samples. First an image from the training set is augmented into two versions (for example, a flipped version and a blurred version). Each version is passed through a separate encoder: the query encoder and key encoder. Both output vector representations of their respective images. The output of the key encoder is added to a dynamically-sized dictionary which stores negative samples with respect to the output of the query encoder. The contrastive loss is then calculated between the query encoder’s output and the vectors in the dictionary. Over time, this process allows the model to learn effective representations of the images in the dataset.

\subsection{SSP in Medical Imaging}

SSP has achieved success in numerous areas, including medical imaging~\cite{4}. Assembling large-scale medical imaging datasets requires domain knowledge, is costly, and is time-consuming, limiting the efficacy of medical imaging models \cite{15}\cite{6}.



\section{Approach}
Our goal is to explore the effect of changing the rotation angle $\theta$ on the learned features when using SSP with the rotation augmentation.

For each rotation angle $\theta$ used in generating the image pairs for contrastive pre-training, we visualize the salient pixels in the image using SmoothGrad~\cite{22}, and compare the salient pixel mask to the ground truth segmentation mask for that image using the Dice coefficient.

We acknowledge that this is an indirect method of observing the effect of setting the rotation angle $\theta$. In future work, more direct methods, such as measuring the performance on a downstream task or visualizing the learned features, will be used to observe the effect of changing the rotation angle $\theta$.

\section{Datasets}

\begin{figure}[h]
    \centering
    \includegraphics[width=1\linewidth]{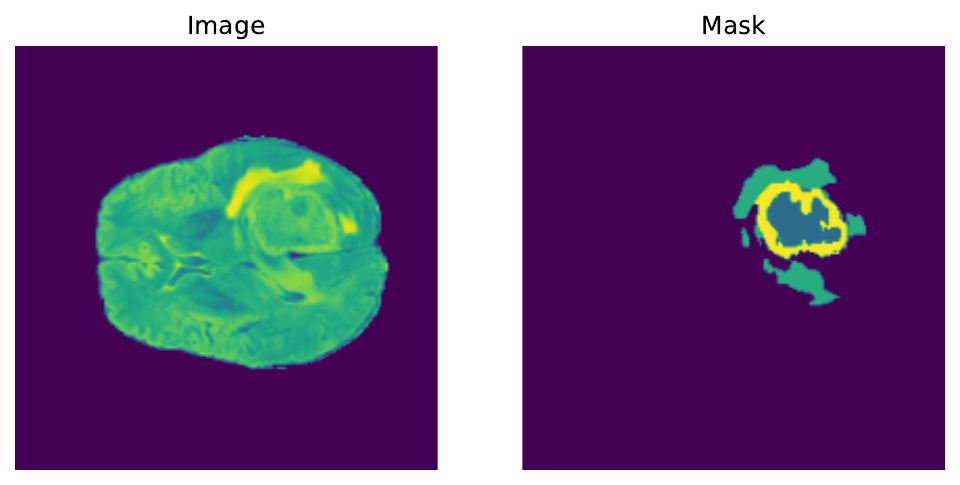}
    \caption{Sample image and segmentation mask from the BraTS dataset}
    \label{fig:brats-ex}
\end{figure}

\begin{figure}[h]
    \centering
    \includegraphics[width=1\linewidth]{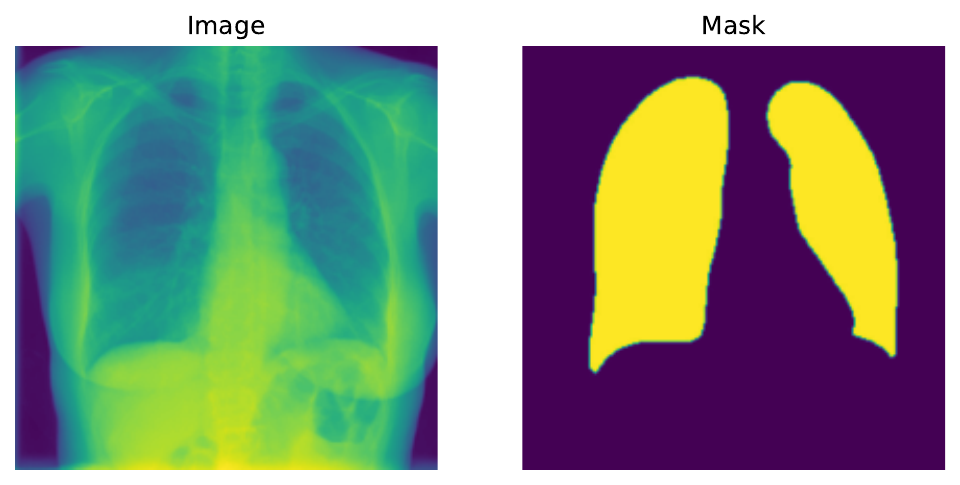}
    \caption{Sample image and segmentation mask from the Lung Mask Image dataset}
    \label{fig:lung-ex}
\end{figure}

\begin{figure}[h]
    \centering
    \includegraphics[width=1\linewidth]{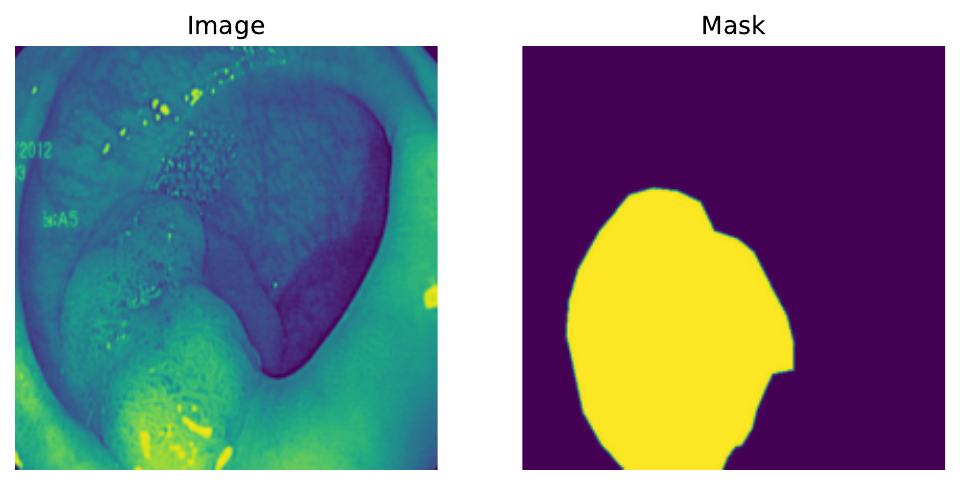}
    \caption{Sample image and segmentation mask from the Kvasir-SEG dataset}
    \label{fig:gi-ex}
\end{figure}

Three publicly available datasets were used in this study:

\begin{itemize}
  \item BraTS: Mutlimodal MRI scans with annotated brain tumour segments (see Fig.~\ref{fig:brats-ex})~\cite{39}.
  \item Lung Mask Image Dataset: Segmented lung X-ray images (see Fig.~\ref{fig:lung-ex})~\cite{43}.
  \item Kvasir-SEG: Gastrointestinal polyp images with corresponding segmentation masks (see Fig.~\ref{fig:gi-ex})~\cite{44}.
\end{itemize}

\subsection{Self-Supervised Model and Fine-Tuning}

In this study, an ImageNet-pre-trained MoCo v2 model with a ResNet-50 backbone was used~\cite{40}. This model was fine-tuned on the medical images of each dataset separately (the segmentation masks were not used at all during training). Fine-tuning was performed for 100 epochs with a batch size of 16. The optimizer used was stochastic gradient descent with a learning rate of 0.01, momentum of 0.9, and weight decay of $1 \times 10^{-4}$.

Prior to training, the images were normalized and subjected randomly to various augmentations to generate the positive pairs, including resizing, Gaussian blurring, horizontal and vertical flips, rotation, and affine translation.

We crop the outer border of the image in order to avoid the network's being able to rely on boundary artifacts to detect rotations by $\theta$.

\subsection{Saliency Map Generation and Evaluation}

\begin{figure}
    \centering
    \includegraphics[width=1\linewidth]{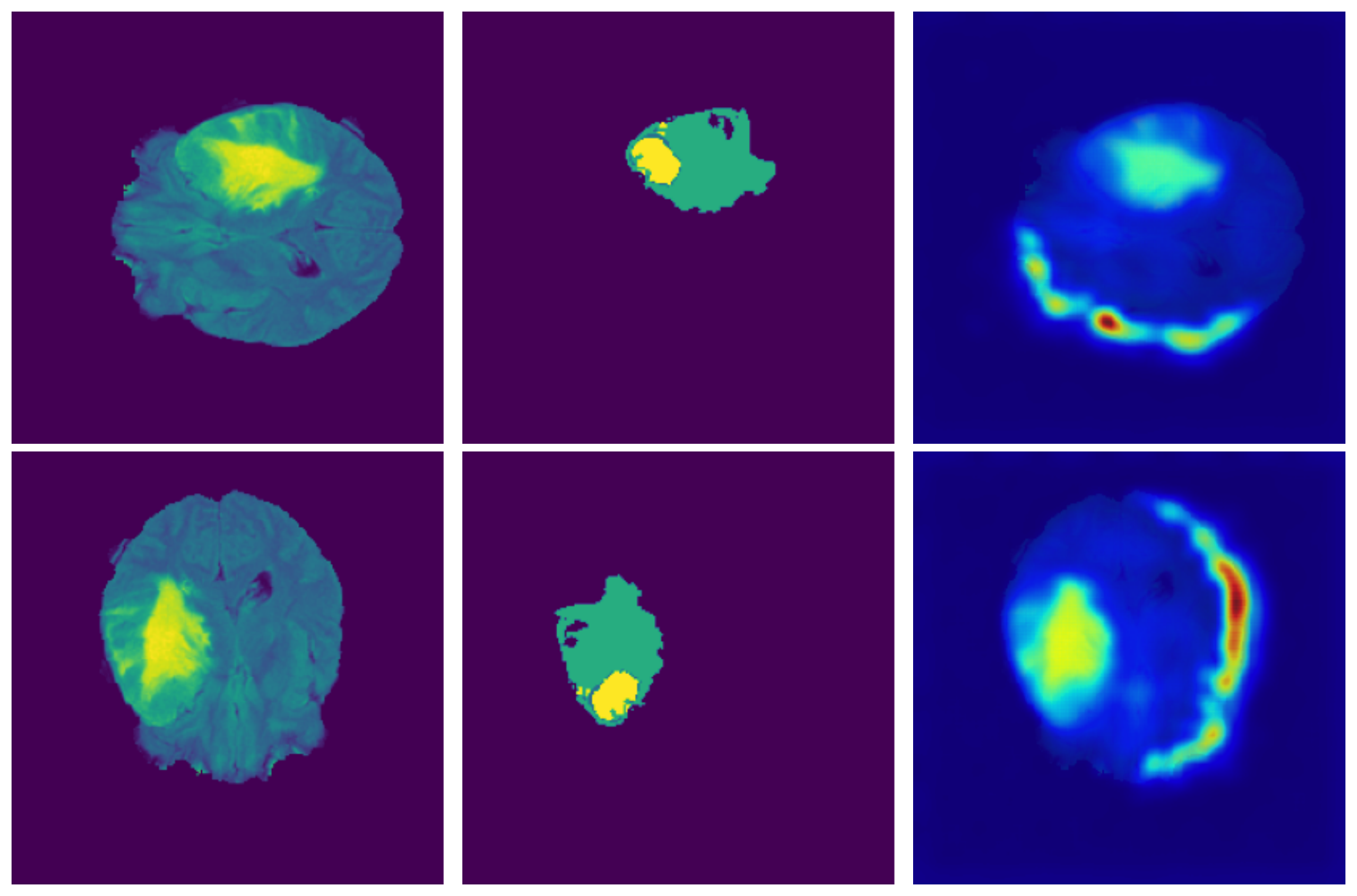}
    \caption{Comparison of original images from Lung Mask Image dataset, segmentation masks, and saliency maps. Top row: Original orientation. Bottom row: Rotated by 95 $\degree$.}
    \label{fig:brats_map}
\end{figure}

\begin{figure}
    \centering
    \includegraphics[width=1\linewidth]{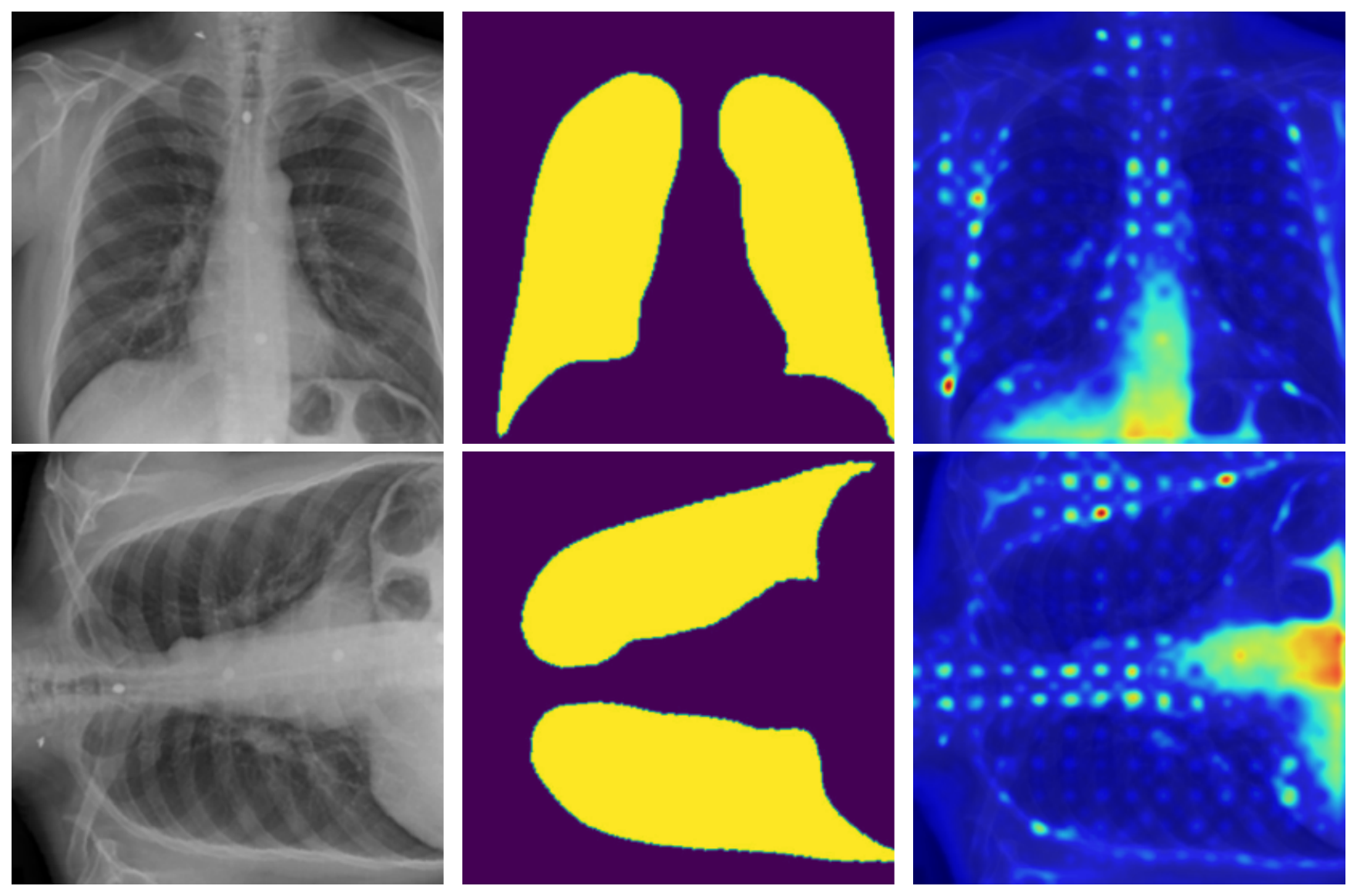}
    \caption{Comparison of original images from Kvasir-SEG dataset, segmentation masks, and saliency maps. Top row: Original orientation. Bottom row: Rotated by 95 $\degree$.}
    \label{fig:lung_map}
\end{figure}

\begin{figure}
    \centering
    \includegraphics[width=1\linewidth]{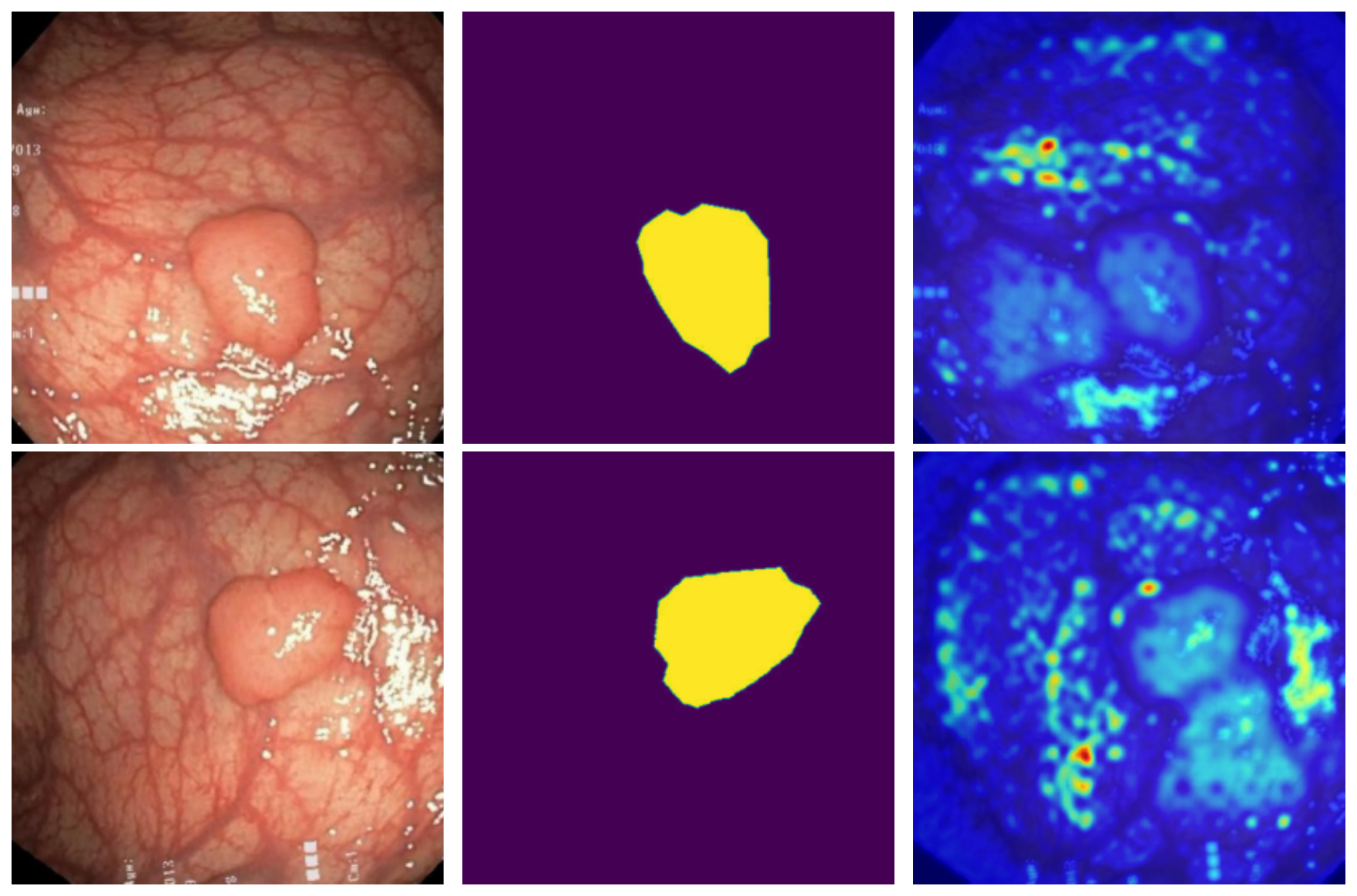}
    \caption{Comparison of original images from BraTS dataset, segmentation masks, and saliency maps. Top row: Original orientation. Bottom row: Rotated by 95 $\degree$.}
    \label{fig:gi-map}
\end{figure}

Following training, to evaluate the regions of the image that were focused on by the model, saliency maps were generated using the SmoothGrad~\cite{22} method, which adds noise to the input images and averages the resulting gradients. See examples in Figs.~\ref{fig:brats_map}~\ref{fig:lung_map}~\ref{fig:gi-map}. The implementation was adapted from a publicly available GitHub repository\footnote{https://github.com/fawazsammani/explain-cl}.

75 images were selected from each dataset and rotated through a full 360-degree range, with saliency maps generated at every single degree increment. The saliency maps were then evaluated based on their similarity to the corresponding ground-truth segmentation masks. The similarity was calculated using the Dice score.

\begin{table*}[!h]

    \centering
    \caption{Average correspondence (using Dice score) between ground truth segmentation and saliency map obtained using trained features vs. rotation angle (left) and accuracy of an SVM classifier classifying original images versus rotated image vs. rotation angle (right).      \label{tab:results}. }
    \begin{tabular}{cc}
        \includegraphics[width=0.45\linewidth]{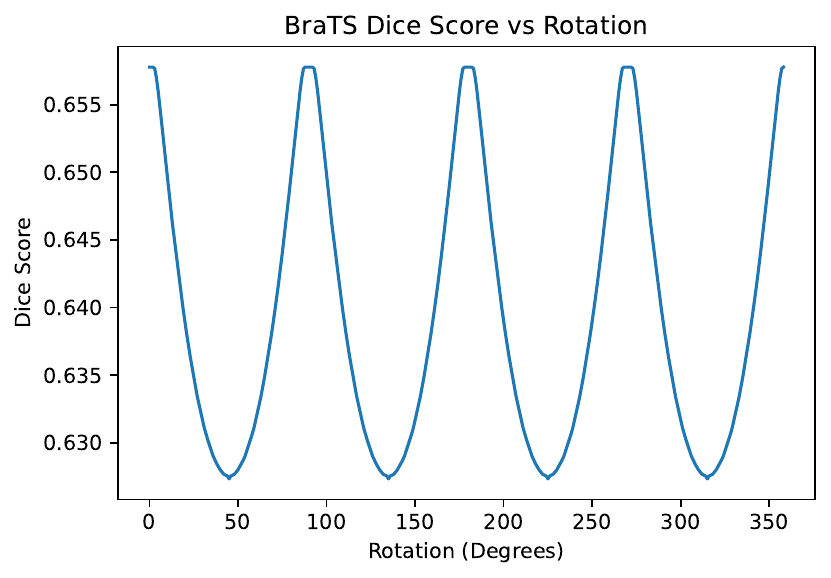} & 
        \includegraphics[width=0.45\linewidth]{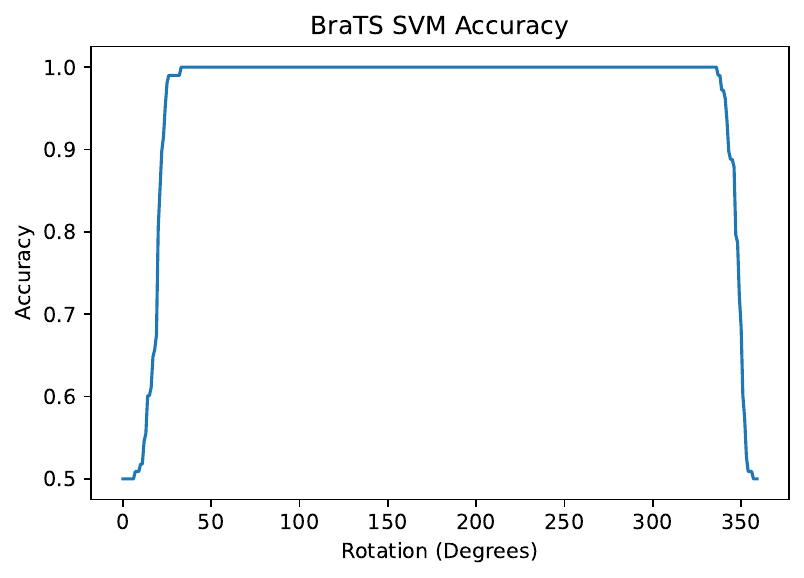} \\
        \includegraphics[width=0.45\linewidth]{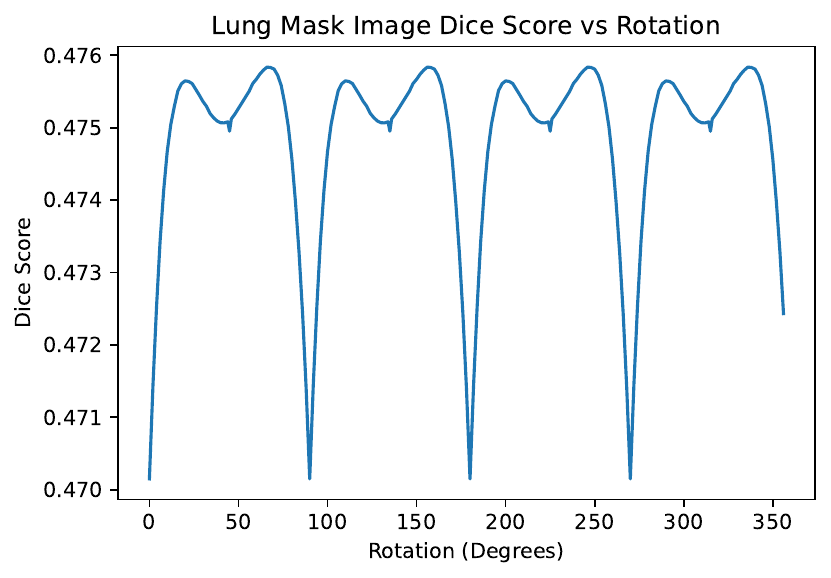} & 
        \includegraphics[width=0.45\linewidth]{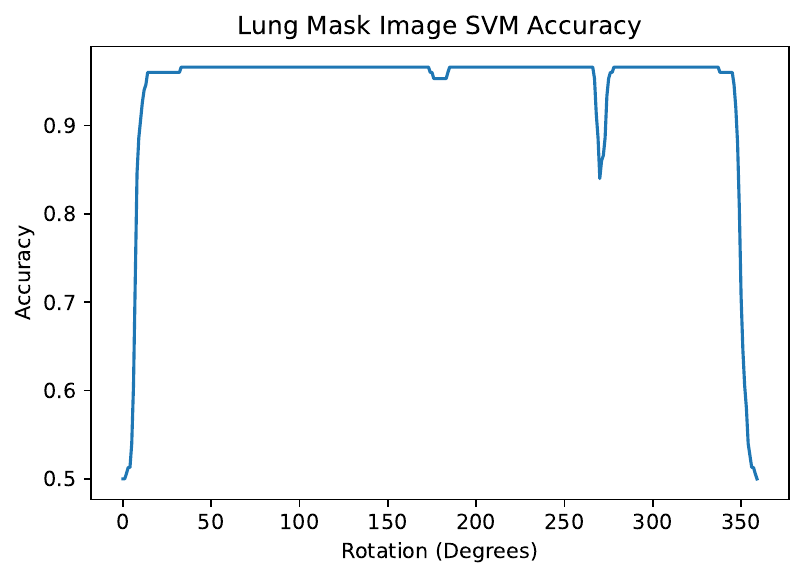} \\
        \includegraphics[width=0.45\linewidth]{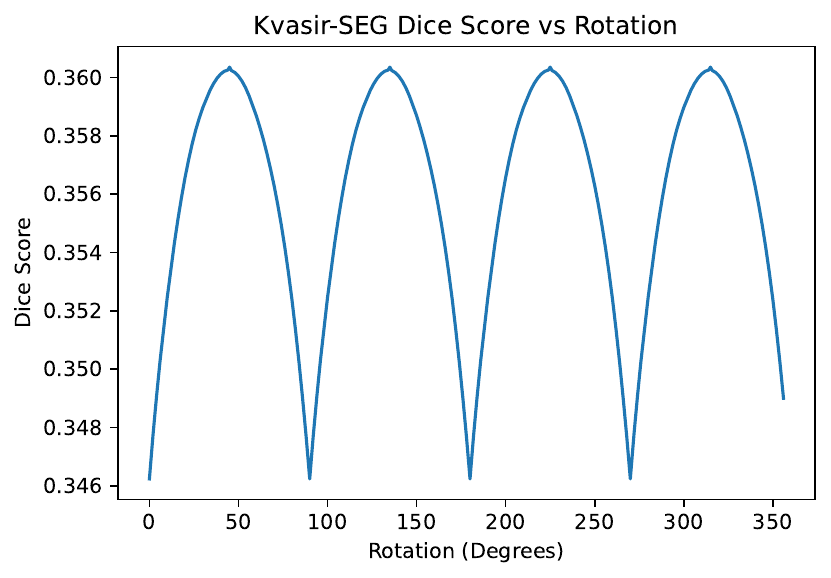} & 
        \includegraphics[width=0.45\linewidth]{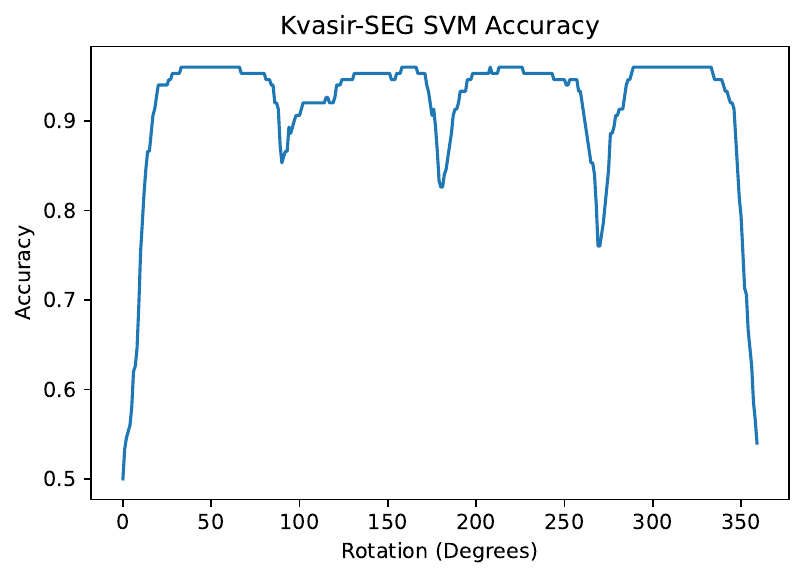} \\
    \end{tabular}
\end{table*}

\section{Results: Saliency Maps and Segmentation Ground Truth}

For each dataset, the Dice score between the 75 saliency maps and their corresponding segmentation masks were averaged for each rotation angle $\theta$. The Dice scores were then plotted against the rotation in degrees to reveal distinctive patterns across each dataset.

The results are in Table~\ref{tab:results}. 

The Dice score versus rotation angle for the BraTS (brain imaging) dataset in exhibits periodic behaviour. Dice scores oscillated between approximately $0.627$ and $0.658$ with peaks around every  $90\degree$ of rotation. The maxima occur at  $90\degree$, $180\degree$, and $270\degree$. The minima occur at  $45\degree$, $135\degree$, $225\degree$, and $315\degree$. 

The plot for the Lung Mask Image dataset reveals another periodic shape with a different amplitude range, namely between approximately $0.470$ and $0.476$. For this dataset the maxima occur at  $66\degree$, $156\degree$, $246\degree$, and $336\degree$ degrees. The minima occur at  $90\degree$, $180\degree$, and $270\degree$.

Another periodic shape is presented in the plot for the Kvasir-SEG (GI tract) dataset, this time oscillating between Dice scores of approximately $0.346$ and $0.360$. The maxima of this plot occur at $45\degree$, $135\degree$, $225\degree$, and $315\degree$. Similarly to the chest dataset, the minima occur at $90\degree$, $180\degree$, and $270\degree$.

\section{Classifying HoG Descriptors of Images}
We observe that angles of $45\degree, 90\degree, 135\degree, ...$ seem to often occur at minima and maxima of the correspondence between the saliency maps and the segmentation ground truth. We hypothesize that, when good features are learned, the saliency maps correspond to the segmentation ground truth better. We hypothesize that worse features are learned when the network can take ``shortcuts" in figuring out the angle $\theta$. For example, the network could rely on Histogram-of-Gradients (HoG)~\cite{lowe1999object}-like features. (Although those features are famously good, they are not specific to our dataset; we do not expect that learning HoG features would be a part of a successful fine-tuning of a network that was already pretrained on ImageNet.)

To explore this hypothesis, we train SVMs to classify the HoG features of images rotated by $\theta$ vs unrotated images. We compute and concatenate HoG descriptors for every $64\times 64$ cell. We use cross-validation to select the best parameters for a Guassian-kernel SVM.

The results are in Table~\ref{tab:results}. We observe that the HoG classification accuracy is low for $\theta$ close to $0\degree$ and (and $360\degree$), reflecting the increased difficulty of the task.

For the Kvasir-SEG dataset, we observe minima in the accuracy of the HoG classifier that correspond to minima in the correspondence between the saliency map and the ground truth segmentation. This seems to be evidence against our theory: when it is more difficult to classify based on HoGs and there are no shortcuts (at least via HoGs), it seems that the correspondence between the saliency map and the ground truth segmentation is lower.

\section{Conclusions and Future Work}
We observe an intriguing property of the rotation pretext text in self-supervised pre-training: it seems that the relationship between the rotation angle and the learned features is periodic and non-monotonic in some sense. The relationship between the rotation angle $\theta$ and the Dice score seems dataset-dependent.

We hypothesize that this is related to the ease with which one can use edges to determine what rotation is performed, but our experiments do not seem to confirm that hypothesis.

In the future, we will measure the learned features more directly, both by visualizing them and by measuring the performance using the features on a downstream task.

\begin{figure}[h]
\centering
\begin{subfigure}{0.3\textwidth}
\centering
\includegraphics[width=\linewidth]{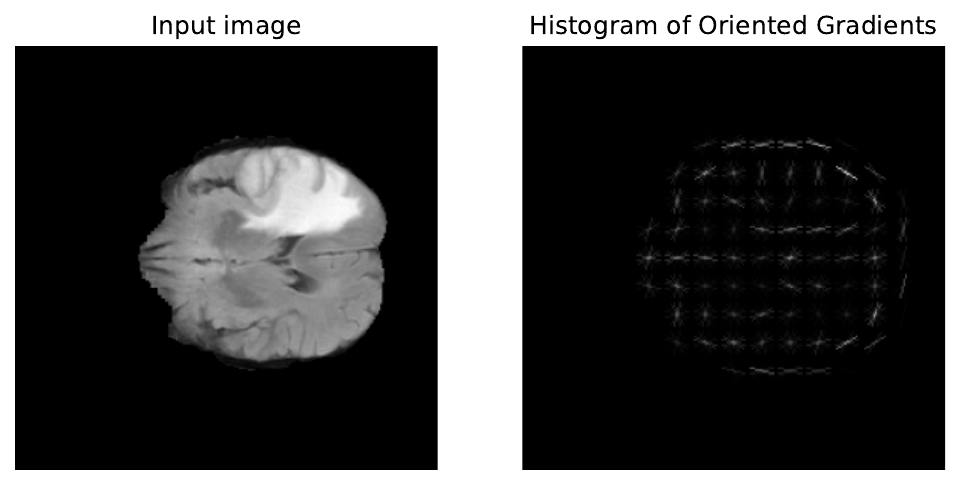}
\caption{Sample HoG features for BraTS dataset}
\label{fig:brats_hog}
\end{subfigure}
\hfill
\begin{subfigure}{0.3\textwidth}
\centering
\includegraphics[width=\linewidth]{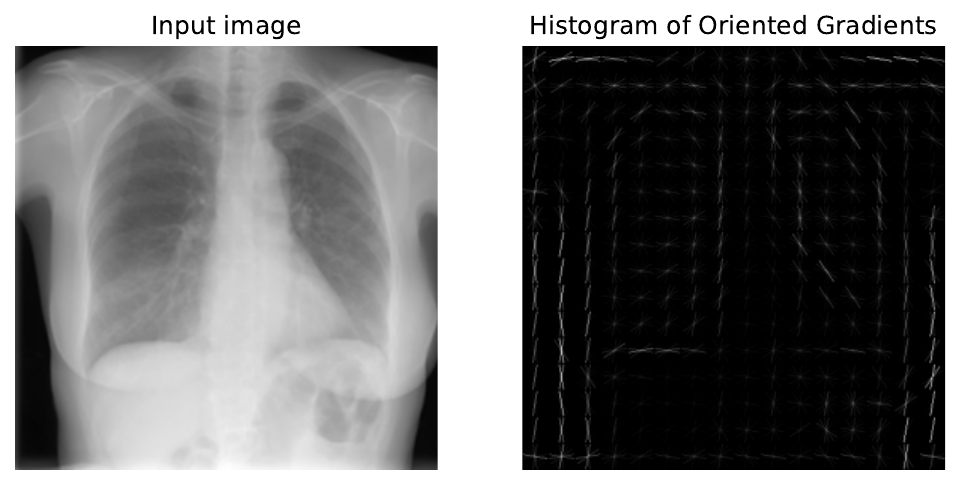}
\caption{Sample HoG features for Lung Mask Imaging dataset}
\label{fig:lung_hog}
\end{subfigure}
\hfill
\begin{subfigure}{0.3\textwidth}
\centering
\includegraphics[width=\linewidth]{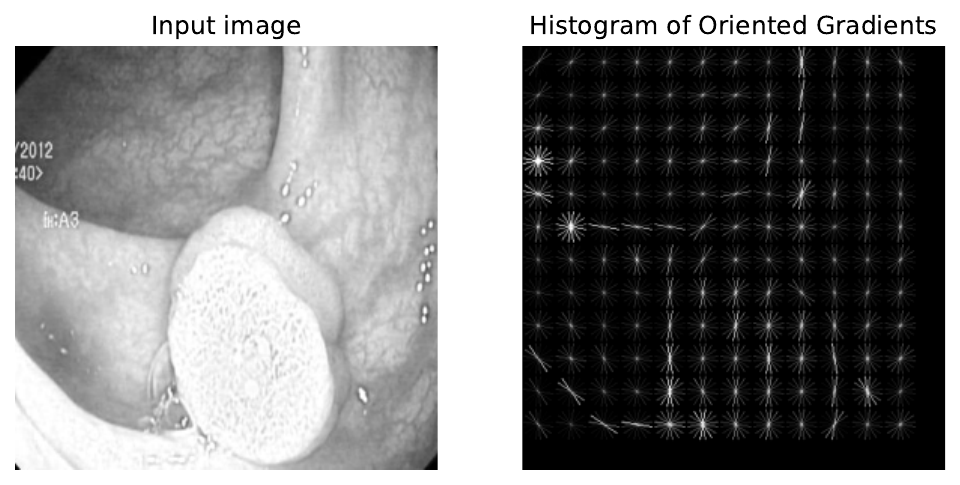}
\caption{Sample HoG features for the Kvasir-SEG }
\label{fig:gi_hog}
\end{subfigure}
\caption{}
\label{fig:combined_map}
\end{figure}

\bibliographystyle{IEEEtran}
\bibliography{IEEEabrv,bibliography}
%

\end{document}